%% file: main.tex
\documentclass[10pt,twocolumn,letterpaper]{article}

\usepackage{wacv}              

\input{preamble}

\definecolor{wacvblue}{rgb}{0.21,0.49,0.74}
\usepackage[pagebackref,breaklinks,colorlinks,allcolors=wacvblue]{hyperref}

\def\wacvPaperID{*****} 
\def\confName{WACV}
\def\confYear{2026}

\title{Efficient Deep Demosaicing with Spatially Downsampled Isotropic Networks}

\author{
Cory Fan\thanks{Work completed principally when Cory Fan was an intern at Omnivision.}\\
Cornell University \\
Ithaca, NY 14853, USA \\
\texttt{cyf5@cornell.edu}
\and
Wenchao Zhang \\
OmniVision Technologies \\
Santa Clara, CA 95054, USA \\
\texttt{wenchao.zhang@ovt.com}
}

\begin{document}
\maketitle
\input{sec/0_abstract}    
\input{sec/1_intro}
\input{sec/2_related}
\input{sec/3_methods}
\input{sec/4_experiments}
\input{sec/5_conclusion}
{
    \small
    \bibliographystyle{ieeenat_fullname}
    \bibliography{main}
}

\end{document}

%% file: preamble.tex
\usepackage{svg}
\usepackage{multirow}
\usepackage{makecell}
\usepackage{pgfplots}
\usepackage{pgfplotstable}
\usepackage{tikz}
\pgfplotsset{compat=1.18}
\usepgfplotslibrary{groupplots}

%% file: sec/0_abstract.tex
\begin{abstract}
In digital imaging, image demosaicing is a crucial first step which recovers the RGB information from a color filter array (CFA). Oftentimes, deep learning is utilized to perform image demosaicing. Given that most modern digital imaging applications occur on mobile platforms, applying deep learning to demosaicing requires lightweight and efficient networks. Isotropic networks, also known as residual-in-residual networks, have been often employed for image demosaicing and joint-demosaicing-and-denoising (JDD). Most demosaicing isotropic networks avoid spatial downsampling entirely, and thus are often prohibitively expensive computationally for mobile applications. Contrary to previous isotropic network designs, this paper claims that spatial downsampling to a signficant degree can improve the efficiency and performance of isotropic networks. To validate this claim, we design simple fully convolutional networks with and without downsampling using a mathematical architecture design technique adapted from DeepMAD, and find that downsampling improves empirical performance. Additionally, empirical testing of the downsampled variant, JD3Net, of our fully convolutional networks reveals strong empirical performance on a variety of image demosaicing and JDD tasks. 
\end{abstract}

%% file: sec/1_intro.tex
\section{Introduction}

\input{sec/fig_1.tex}

Demosaicing is a first and crucial step in an Image Sensor Processing (ISP) pipeline \cite{Karaimer}. Recent advances in mobile photography, including more challenging non-Bayer CFA mosaics and pixel binning techniques, have made the task of demosaicing in modern mobile phones increasingly difficult \cite{ESUM}. Thus, deep learning techniques for non-Bayer demosaicing have been increasingly explored \cite{BJDD,SAGAN,ESUM}. However, deep learning techniques are infamously compute-heavy, which poses challenges for usage on a mobile platform. Consequently, lightweight networks are essential for demosaicing on mobile phones \cite{BMTNet}.

One major class of image demosaicing network are isotropic networks, which are often called residual-in-residual networks \cite{JDNDMSR,ESUM,SGNet}. These networks are characterized by their entire trunk maintaining the same spatial resolution; one particularly well-known example is the Vision Transformer (ViT) architecture\cite{ViT}. However, unlike ViT networks, which perform aggressive patch-ification to reduce computational cost, isotropic networks for demosaicing largely perform their computations on the full spatial resolution of the image (or nearly so), consequently leading to a comparatively higher computational cost. This particular property of isotropic image demosaicing networks seems to be inherited from the strong success of isotropic networks in super-resolution\cite{RCAN, SwinIR, X-Restormer}. On the contrary, for the task of image demosaicing, we find that spatial downsampling is effective for improving performance and efficiency in FLOP-equivalent network variants.

Reasoning about the effect of singular variables in designing deep networks can often be challenging, due to the dozens of interdependent hyperparameters found in network architectures. We employ two measures to enable empirical testing of the effects of downsampling. First, we consider only very simple fully-convolutional isotropic networks, which reduces our search space to only three variables: depth, width, and downsampling factor. Next, to produce principled and fair candidate variants of downsampled and non-downsampled networks, we employ a mathematical architecture design technique modified from DeepMAD\cite{DeepMAD}. Our results show that downsampling improves the performance of two FLOP-equivalent networks, especially in particularly small networks. 

\begin{figure}[ht!] 
    {\fontsize{6}{8}\selectfont
        \centering 
    \includegraphics[width=0.47\textwidth]{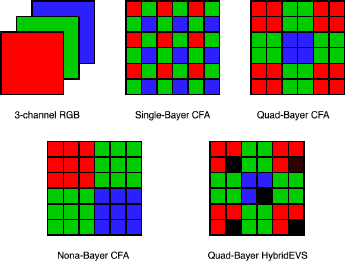}
    }
    \caption{CFAs investigated in this paper. Networks often have to deal with challenging CFAs with missing information (HybridEVS CFA, for instance) or multiple CFA simultaneously.}
    \label{fig:cfas} %
\end{figure}

Recent work in image demosaicing has explored complex patterns\cite{BJDD, SAGAN, BMTNet} and techniques to create unified models\cite{KLAP, ESUM}. Following this diversity in recent image demosaicing work, we test the applicability of downsampled networks to a variety of tasks. Our downsampled networks, which we term JD3Net (\textbf{J}oint \textbf{D}emosaicing and \textbf{D}enoising with \textbf{D}ownsampled \textbf{Net}work), are highly simple fully convolutional networks, but nonetheless we find that they achieve state-of-the-art efficiency on unified Bayer/non-Bayer joint-demosaicing-and-denoising, classical Bayer image demosaicing, and quad-Bayer HybridEVS image demosaicing (see \cref{fig:1}). An overview of the CFA patterns considered can be found in \cref{fig:cfas}. Overall, we argue that downsampling can broadly improve the efficiency of deep networks across a variety of image demosaicing tasks.

%% file: sec/fig_1.tex
\begin{figure}[t]
\centering
\begin{tikzpicture}
\begin{groupplot}[
    group style={
        group size=1 by 2,
        vertical sep=1.2cm
    },
    width=\columnwidth,
    height=0.65\columnwidth,
    xmode=log,
    log basis x=10,
    xlabel={FLOPs (G) in log scale},
    grid=both,
    grid style={dashed, gray!30},
    tick label style={font=\footnotesize},
    label style={font=\footnotesize},
]

\nextgroupplot[
    ylabel={Avg. PSNR - HDD Unified},
    ymin=50, ymax=51.3,
    xmin=15, xmax=600,
]

\addplot[
    color=red,
    mark=diamond*,
    line width=1.4pt
] coordinates {
    (25, 50.60)
    (128, 51.05)
};
\node[red, rotate=25, font=\footnotesize] 
    at (axis cs:60,50.93) {JD3Net (ours)};

\addplot[color=black, mark=*] coordinates {(106, 50.08)};
\node[font=\footnotesize] at (axis cs:110,50.15) {KLAP};

\addplot[color=brown, mark=*] coordinates {(408, 50.60)};
\node[font=\footnotesize] at (axis cs:400,50.7) {ESUM};

\nextgroupplot[
    ylabel={Avg. PSNR - HybridEVS Quad-Bayer},
    ymin=36.3, ymax=37.8,
    xmin=20, xmax=700,
]

\addplot[
    color=red,
    mark=diamond*,
    line width=1.4pt
] coordinates {
    (25, 36.734)
    (128, 37.648)
};
\node[red, rotate=37, font=\footnotesize] 
    at (axis cs:60,37.3) {JD3Net (ours)};

\addplot[
    color=black,
    mark=*,
    line width=1.4pt
] coordinates {
    (37, 36.6575)
    (149, 37.47)
};
\node[black, rotate=37, font=\footnotesize] 
    at (axis cs:80,37) {TSANet};


\addplot[color=gray, mark=square*] coordinates {(32, 36.554)};
\node[font=\footnotesize] at (axis cs:34,36.45) {NAFNet(w-32)};

\addplot[color=purple, mark=*] coordinates {(282, 36.834)};
\node[font=\footnotesize] at (axis cs:282,36.94) {Restormer};

\addplot[color=gray!70, mark=*] coordinates {(491, 37.47)};
\node[font=\footnotesize] at (axis cs:400,37.6) {DemosaicFormer};

\end{groupplot}
\end{tikzpicture}

\caption{PSNR vs. computational cost on unified joint-demosaicing-and-denoising on HDD (top) and Quad-Bayer HybridEVS demosaicing (bottom).}
\label{fig:1}
\end{figure}

%% file: sec/2_related.tex
\section {Related Work}

\subsection{Tasks in Demosaicing}

Modern methods for classical Bayer demosaicing, including deep learning methods, can reconstruct RGB images with a high-degree of precision. Nonetheless, numerous challenges still exist in the field of demosaicing, including joint-demosaicing-and-denoising and non-Bayer demosaicing techniques.

Joint demosaicing and denoising (JDD) is a deep learning task that combines two image processing steps – demosaicing and denoising – into one end-to-end objective. JDD is attractive because it preserves the decorrelated noise of raw images while maintaining the full information of the raw image for demosaicing\cite{DeepJoint}; consequently, research in this direction has existed even before the popularization of deep learning \cite{Klatzer}. Nonetheless, deep learning approaches have been the mainstream in JDD research recently. Typically, a convolutional neural network (CNN) variant is applied to this task \cite{SGNet,BJDD,SAGAN,ESUM,JDNDMSR}, with some exceptions \cite{DemosaicFormer,BMTNet}.

While early JDD research was largely based on the traditional Bayer CFA \cite{JDNDMSR,SGNet,TENet}, modern research often focuses on non-Bayer demosaicing\cite{BJDD,SAGAN,TSANet,BMTNet} or unified Bayer/non-Bayer demosaicing\cite{KLAP,ESUM}. Non-Bayer CFAs tend to be more challenging, and can even have pixels with completely zero information, such as in HybridEVS image demosaicing. The difficulty of non-Bayer demosaicing, even without the denoising task, similarly merits deep learning.

\subsection {Isotropic Networks for Demosaicing}
The neural networks applied to demosaicing can mainly be subdivided into two types of macro-architectures: U-Net \cite{BJDD,SAGAN,DemosaicFormer,BMTNet,KLAP,TSANet} and Isotropic (often called residual-in-residual)\cite{SGNet,JDNDMSR,ESUM,DeepJoint,RNAN,GRL}. 

In general, the main difference between U-Net and isotropic macro-architectures lies in downsampling. U-Net networks employ progressive downsampling to capture multiple feature-scales \cite{UNet}, while the trunks of isotropic networks employ a uniform spatial size \cite{ESUM,JDNDMSR,RCAN}, and employ spatial downsampling or upsampling only at the beginning or end of the network.

For most general image restoration tasks, such as image denoising or super-resolution, an isotropic network employs no downsampling at all \cite{RCAN,SwinIR}. For demosaicing, while many isotropic networks also do not perform downsampling \cite{ESUM, GRL, RNAN}, some networks perform "packing", which folds a CFA pattern into a single pixel\cite{DeepJoint,JDNDMSR}. However, prior networks do not explicitly use downsampling to improve efficiency and do not exceed 2x downsampling. A brief summary of prior isotropic networks for demosaicing is shown in \cref{table:taxonomy}. 

\begin{table}
\small	
\setlength{\tabcolsep}{5pt}
\renewcommand{\arraystretch}{1.2}
{\centering
\begin{tabular}{c|c|c|c}
\hline
Name & Year & Downsampling Ratio & GFLOPs \\
\hline
DeepJoint\cite{DeepJoint} & 2016 & 2 & 14 \\
MMNet\cite{MMNet} & 2018 & 1 & 18 \\
RNAN \cite{RNAN} & 2019 & 1 & 162 \\
TENet \cite{TENet} & 2019 & 2 & 3199 \\
SGNet \cite{SGNet} & 2020 & 2 & - \\
JDNDM\cite{JDNDMSR} & 2021 & 2 & 408 \\
GRL\cite{GRL} & 2023 & 1 & 491 \\
ESUM\cite{ESUM} & 2025 & 1 & 408 \\
\hline
\end{tabular}
\par}
\caption{Previous Isotropic Networks Applied to Demosaicing. FLOPs calculated for 256x256 image. SGNet does not have fully available code, so FLOPs cannot be calculated. Previous isotropic demosaicing networks do not use downsampling greater than two.} 
\label{table:taxonomy}
\end{table}

Because isotropic networks tend to operate on high spatial resolutions, without employing spatial downsampling, they are often quite slow and inefficient\cite{X-Restormer}. This is a notable problem for demosaicing, since its main application field is mobile phones, which are famously compute constrained. In particular, isotropic networks for JDD can be several hundred G-FLOPs for 256-by-256 pixel images\cite{ESUM,JDNDMSR}, which are several orders of magnitude larger than networks typically designed for image recognition on mobile devices\cite{MBv2,MBv3}.

In this paper, we argue that a crucial step in improving the efficiency of an isotropic JDD network is deliberately employing spatial downsampling for the express purpose of improving efficiency, at a factor greater than previously explored.

\subsection{Zero-Shot Neural Architecture Search}
The primary tool we will use to make this argument is Zero-Shot Neural Architecture Search (NAS). In contrast to conventional NAS techniques, Zero-Shot NAS attempts to find a performant neural network architecture without any training, often based on mathematical observations regarding the nature of deep neural networks\cite{DeepMAD,Zen-NAS,MAE-Det}. This is attractive for designing networks in a principled manner, and is the primary reason we employ Zero-Shot NAS techniques in this paper.

Often, Zero-Shot NAS techniques employ the Principle of Maximum Entropy, which aims to maximize the entropy of the feature map outputted by the network \cite{DeepMAD,Zen-NAS}. In particular, this paper focuses on the DeepMAD\cite{DeepMAD} entropy formulation, which is designed for convolutional neural networks and takes the following form:

\begin{equation}
H_L \triangleq \log (r_{L+1}^2c_{L+1})\sum_{i=1}^L\log(c_ik_i^2/g_i).
\end{equation}

In the above equation, $H_L$ is the entropy, $r_{L+1}$ is the final resolution (before global average pooling), $c_{L+1}$ is the final number of channels, $c_i$, $k_i$ and $g_i$ are the input channel, kernel size, and number of convolutional groups to the $i$-th layer respectively.

To use this entropy score to find performant CNNs, DeepMAD constrains the ratio of width to depth $\rho = L/w$, as well as the desired FLOPs and parameters, and then finds the CNN with maximum entropy under the constraints. Thus, DeepMAD compresses a complex search with many parameters into a simple search across a single parameter $\rho$, which can be done via mathematical programming. We will employ a modified version of this technique to compare downsampled and non-downsampled isotropic networks.

%% file: sec/3_methods.tex
\section{Methods}

\subsection{Network Design}

Our primary goal in this paper is to compare downsampled and non-downsampled isotropic networks. To do this, we make the following design choices:
\begin{enumerate}
\item First, we design JD3Net as a simple fully convolutional network, forgoing any complex attention mechanisms that can be often found in image restoration. By using highly simple networks, we can be more confident that results are attributable to macro-architecture design.
\item Second, we employ mathematical architecture design to choose our specific network parameters, such as width or downsampling ratio. Using mathematical architecture design allows us to design representative architectures for downsampled and non-downsampled variants of JD3Net in a principled manner, without searching across the entire space of potential JD3Net networks.
\end{enumerate} 

\begin{figure}[ht!] 
    \centering 
    \quad
    \includegraphics[width=0.5\textwidth]{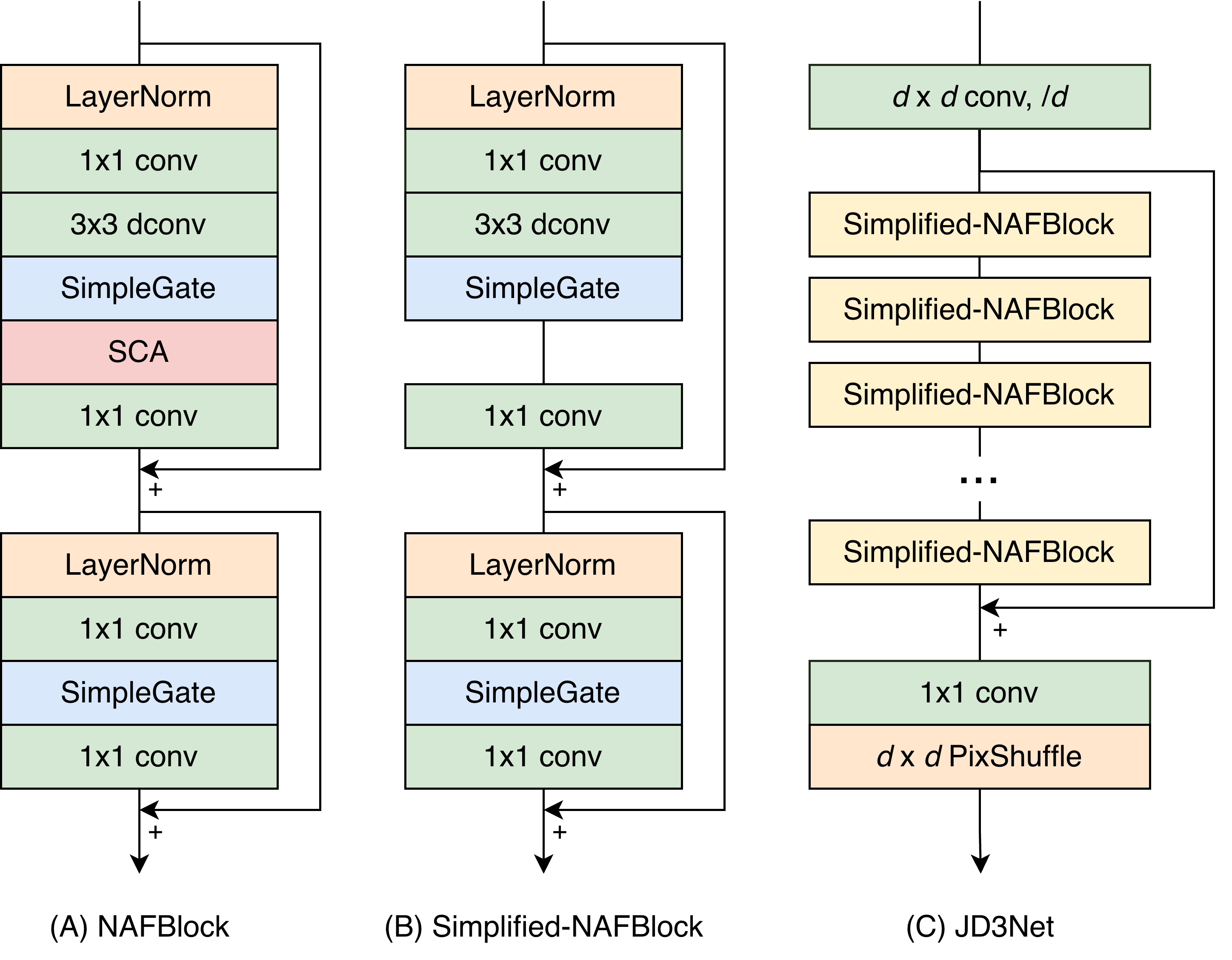} 
    \caption{Simplified-NAFBlock and JD3Net Architecture. (A) NAFBlock. SCA stands for Simple Channel Attention. Dconv stands for depthwise convolution. (B) Simplified-NAFBlock, which is the same as NAFBlock except for removal of SCA. (C) Our fully-convolutional JD3Net architecture. JD3Net is fully-convolutional and highly simple.}
    \label{fig:design} %
\end{figure}

JD3Net employs modified variants of NAFBlocks\cite{NAFNet}, with the attention removed, and which we will refer to as Simplified-NAFBlocks. A specific illustration of NAFBlocks and Simplified-NAFBlocks can be found in \cref{fig:design}. 

Our decision to remove channel attention from JD3Net is due to the implementation complexity of attention in image restoration networks. For instance, techniques such as Test-time Local Converter\cite{TLC} are used by some networks, but not others. Moreover, we provide some brief ablation in \cref{sec:ablation}, which shows that the Simple Channel Attention (SCA) mechanism used by NAFNet actually decreases the performance of JD3Net. Since JD3Net can acheive strong empirical performance without complex attention mechanisms, we leave the investigation of attention for downsampled isotropic networks to future work.

In terms of the overall architecture of JD3Net, we employ a typical isotropic design. Where $d$ is the downsampling ratio and $B$ is the number of blocks, our networks consist of a $d$x$d$ convolutional layer to begin, then $B$ Simplified-NAFBlocks, and finally a 1x1 convolution and a $d$x$d$ PixShuffle layer to return our image to the original resolution (see \cref{fig:design}).

We detail our selection of depth, width, and downsampling ratio with mathematical architecture design in the following section.

\subsection{Mathematical Architecture Design}

As previously stated, we modify DeepMAD's entropy score to design JD3Net. The principal issue with applying DeepMAD directly to image restoraiton is that its entropy score is reliant on the spatial resolution of the input image (ie a network would have a different entropy score with a 256x256 image compared to a 512x512 image). Consequently, DeepMAD selects different networks depending on the input resolution. For classification, where input resolution is an active consideration in network design \cite{EfficientNet,ConvNext}, this makes sense. But for image restoration, where networks should be able to process an arbitrary image, we would prefer our selection of network to be invariant to the input resolution. 

To achieve a resolution-invariant entropy score, instead of summing across the size of the entire final feature map, we utilize the channel density (number of channels per output pixel). Our modified entropy score is thus as such:

\begin{equation}
H_L \triangleq \log (c_{L+1}/d^2)\sum_{i=1}^L\log(c_ik_i^2/g_i).
\end{equation}

Similar to DeepMAD, we perform a search across the ratio of width and depth $\rho$. However, noting the heuristical nature of entropy scores, we also find that our entropy score over-promotes downsampling, with downsampling ratios of 6 or 8 in larger networks. However, since a strong local modeling capability is still crucial to the success of isotropic networks, we find more success constraining the downsampling ratio to 4. Additionally, unlike DeepMAD, since our primary concern is FLOPs efficiency, we do not apply a parameter constraint. Finally, since the number of blocks is linearly related with the number of convolutions, for notational simplicity, we constrain $\rho = B/w$, where B is the number of blocks. 

Thus, our approach takes the following mathematical programming formulation:

\begin{equation}
\begin{aligned}
\max_{w, d, B}\quad& H_L \\ 
\text{s.t.}\quad& B/w \leq \rho \\
\quad & d\leq 4 \\
& \text{FLOPs} \leq \text{budget}
\end{aligned}
\end{equation}

Solving this mathematical programming problem will result in the entropy-maximizing values for $B$, $w$, and $d$, under the constraint of $\rho$. In \cref{sec:search}, we will show that the optimal downsampling ratio $d$ is greater than one for most practical choices of network, and perform our search to find an effective network. 

%% file: sec/4_experiments.tex
\section{Experiments and Analysis}

\subsection{Searching for JD3Net}
\label{sec:search}

\begin{table}[b]
\small	
\setlength{\tabcolsep}{5pt}
\renewcommand{\arraystretch}{1}
{\centering
\begin{tabular}{c|c|c|c|c|c|c}
\hline
Network & \makecell{GFLOPs\\Constraint} & $\rho$ & $d$ & $w$ & $B$ & Val. PSNR \\
\hline
& & 0.5 & 4 & 96 & 48 & 44.146 \\
& & 0.7 & 4 & 88 & 60 & 44.132 \\
JD3Net-S & 25 & \textbf{1.0} & \textbf{3} & \textbf{64} & \textbf{64} & \textbf{44.148 }\\
& & 1.2 & 3 & 60 & 72 & 44.146 \\
& & 1.5 & 3 & 56 & 84 & 44.136 \\
\hline
& & 0.5 & 1 & 40 & 16 & 44.02 \\
& & \textbf{0.7} & \textbf{1} & \textbf{36} & \textbf{20} & \textbf{44.03} \\
JD3Net-S-x1 & 25 & 1.0 & 1 & 28 & 28 & 43.99 \\
& & 1.2 & 1 & 28 & 32 & 44.02 \\
& & 1.5 & 1 & 24 & 36 & 43.89 \\
\hline
&  & 0.5 & 4 & 172 & 84 & 44.136 \\
&  & 0.7 & 4 & 152 & 104 & 44.286 \\
JD3Net & 128 & 1.0 & 4 & 136 & 132 & 44.272 \\
&  &\textbf{1.2}&\textbf{4} & \textbf{128} & \textbf{152} & \textbf{44.294} \\
&  & 1.5 & 4 & 120 & 172 & 44.284 \\
\hline
&  & 0.5 & 1 & 68 & 32 & 44.209 \\
&  & 0.7 & 1 & 60 & 40 & 44.216 \\
JD3Net-x1 & 128 & 1.0 & 1 & 52 & 52 & 44.242 \\
&  &1.2& 1 & 48 & 56 & 44.232 \\
&  & \textbf{1.5} & \textbf{1} & \textbf{48} & \textbf{64} & \textbf{44.244} \\
\hline
\end{tabular}
\par}
\caption{Search Results for JD3Net and non-downsampled JD3Net. FLOPs on a 256x256 pixel image.} 
\label{table:search}
\end{table}

As noted previously, we design two networks, one smaller and one larger. JD3Net was constrained to 128 GFLOPs for a 256x256 image, which is more comparable to existing image restoration and JDD networks, while JD3Net-S was constrained to only 25 GFLOPs for a 256x256 image, which is more suitable for mobile applications. For simplicity, we let $w$ and $B$ be multiples of 4. 

We perform our search for JD3Net by comparing validation PSNRs on the Hard Demosaicing Dataset\cite{ESUM}, a dataset of 638 real world joint-demosaicing-and-denoising images from 17 difficult scenes. Specifically, we compare results on the most noisy setting, ISO 3200. We train for 150 epochs using the provided pre-processed 48x48 hard training patches. Learning rate is $0.001$ on Adam optimizer with MSE loss (following ESUM\cite{ESUM}). Batch size for the smaller network is 64, and 48 for the larger network.

For JD3Net-S, we find that $\rho=1.0$ is optimal, resulting in $(d,w,B) = (3,64,64)$, while for base JD3Net, $\rho=1.2$ is optimal, resulting in $(d,w,B) = (4,128,152)$. Thus, our search results reveal that $d>1$ maximizes entropy for most practical isotropic networks.

Additionally, we perform a similar search for networks without downsampling. We refer to the small and normal variants as JD3Net-S-x1 and JD3Net-x1. All search results can be found in \cref{table:search}

\begin{table}
\small	
\setlength{\tabcolsep}{5pt}
\renewcommand{\arraystretch}{1.5}
{\centering
\begin{tabular}{c|c|c|c}
\hline
Network & Single & Quad & Nona \\
\hline
JD3Net-S-x1 & 48.83 & 48.37 & 47.86 \\
\textbf{JD3Net-S} & \textbf{49.06} & \textbf{48.61} & \textbf{48.26} \\
\hline
JD3Net-x1 & 49.27 & 48.85 & 48.43 \\
\textbf{JD3Net} & \textbf{49.36} & \textbf{48.99} & \textbf{48.54} \\
\hline
\end{tabular}
\caption{Comparison of Downsampled and non-Downsampled JD3Nets on HDD ISO3200. Metric is Test PSNR. } 
\label{table:search_test}
\par}
\end{table}

To compare these networks, we assess on the test set of the ISO3200 split of the HDD dataset, shown in \cref{table:search_test}. As expected, JD3Net and JD3Net-S with downsampling outperform their counterparts without downsampling. JD3Net-S, the smaller network of the two, in particular experiences a large improvement in performance with downsampling, with a 0.29 PSNR increase on average. Overall, these results indicate the efficacy of downsampling in designing efficient isotropic demosaicing networks.

\begin{table*}[!t]
\small
\setlength{\tabcolsep}{4pt}
\renewcommand{\arraystretch}{1.5}
{\centering%
\begin{tabular}{c|c|c|c|c|c|c|c|c|c|c|c|c|c|c}
\hline
 \multirow{2}{3.5em}{Network} & \multirow{2}{3.5em}{GFLOPs}& \multicolumn{3}{|c|}{ISO 400} & \multicolumn{3}{|c|}{ISO 800} & \multicolumn{3}{|c|}{ISO 1600} & \multicolumn{3}{|c|}{ISO 3200} &  \multirow{2}{2em}{Avg} \\
 \cline{3-14}
 & & Single & Quad & Nona & Single & Quad & Nona & Single & Quad & Nona & Single & Quad & Nona \\
 \hline
 JDNDM\cite{JDNDMSR} & 408 & 53.69 & - & - & \color{blue} 52.34 & - & - & 50.43 & - & - & 49.00 & - & - & - \\
 BJDD\cite{BJDD} & 69 & - & 50.86 & - & - & 50.05 & - & - & 48.88 & - & - & 47.50 & - & - \\
 SAGAN\cite{SAGAN} & 342 & - & - & 49.55 & - & - & 49.06 & - & - & 48.05 & - & - & 46.88 & - \\
 KLAP\cite{KLAP} & 106 & 53.27 & 52.01 & 50.44 & 51.91 & 50.99 & 49.79 & 50.28 & 49.40 & 48.59 & 48.95 & 48.10 & 47.20 & 50.08 \\
 ESUM\cite{ESUM} & 408 & \color{blue} 53.75 & \color{blue} 52.68 & \color{blue} 51.96 & 52.17 & \color{blue} 51.36 & 50.76 & \color{blue} 50.64 & \color{blue} 50.01 & 49.46 & 48.98 & 48.57 & 48.11 & \color{blue}50.70 \\

 \hline

 JD3Net-S & 25 & 53.42 & 52.34 & 50.92 & 52.21 & 51.35 & \color{blue} 50.92 & 50.56 & 49.95 & \color{blue} 49.56 & \color{blue} 49.06 & \color{blue} 48.61 & \color{blue} 48.26 & 50.60 \\
 
 JD3Net & 128 & \textbf{54.06} & \textbf{53.16} & \textbf{52.38} & \textbf{52.64} & \textbf{51.95} & \textbf{51.29} & \textbf{50.92} & \textbf{49.42} & \textbf{49.86} & \textbf{49.36} & \textbf{48.99} & \textbf{48.54} & \textbf{51.05} \\
 \hline
\end{tabular}%
\par}
\caption{HDD Results (PSNR). Bold is best. Blue is second-best. FLOPs on 256x256 pixel image. Average PSNR is calculated for unified models.} 
\label{table:HDD_results}
\end{table*}

\begin{table*}[!t]
\small	
\setlength{\tabcolsep}{4pt}
\renewcommand{\arraystretch}{1.5}
{\centering
\begin{tabular}{c|c|c|c|c|c|c|c}
\hline
Network & GFLOPs & Kodak & McMaster & BSD100 & Urban100 & WED
 & Avg \\ \hline
BMTNet*\cite{BMTNet} & 7 & 37.69/0.980 & 34.79/0.950 & 36.11/0.981 & 34.45/0.973 & 33.95/0.965 & 35.398/0.970 \\
NAFNet\cite{NAFNet}(w-32) & 32 & 38.60/0.980 & 36.18/0.961 & 37.12/0.985 & 35.63/0.978 & \textbf{35.24}/0.972 & 36.554/0.975 \\
TSANet-S\cite{TSANet} & 37 & 38.73/0.984 & - & 36.56/0.984 & \textbf{36.15}/\color{blue}0.980 & \color{blue}35.19/0.973 & \color{blue}36.658/0.980 \\
JD3Net-S (Ours) & 25 & \textbf{38.93/0.990} & \textbf{36.57/0.985} & \textbf{37.18/0.987} & \color{blue}35.90\color{black}/\textbf{0.985} & 35.09/\textbf{0.983} & \textbf{36.734/0.986} \\
\hline
Restormer\cite{Restormer} & 282 & 39.16/0.986 & 36.54/0.967 & 37.11/0.985 & 36.36/0.977 & 35.00/0.971 & 36.83/0.977 \\
SAGAN\cite{SAGAN} & 342 & 36.14/0.959 & 32.58/0.939 & 30.53/0.931 & 29.89/0.946 & 28.22/0.917 & 31.47/0.938 \\ 
TSANet-L\cite{TSANet} & 149 & \color{blue}39.40/\color{blue}0.986 & - & 37.34/\color{blue}0.986 & 37.07/\color{blue}0.983 & \color{blue}35.76/\color{black}0.977 & 37.393/\color{blue}0.983 \\
DemosaicFormer\cite{DemosaicFormer} & 491 & 39.32/0.982 & \textbf{37.88}/\color{blue}0.963 & \color{blue}37.65\color{black}/0.982 & \textbf{37.64}/0.980 & 34.86/\color{blue}0.986 & \color{blue}37.47/\color{black}0.979 \\
JD3Net (Ours) & 128 & \textbf{39.63/0.991} & \color{blue}37.40/\color{black}\textbf{0.987} & \textbf{37.85/0.989}& \color{blue}37.18/\color{black}\textbf{0.987} & \textbf{36.18/0.987} & \textbf{37.64/0.988} \\
\hline
\end{tabular}
\par}
\caption{Synthetic Quad-Bayer HybridEVS Demosaicing Results (PSNR/SSIM). Top are smaller networks, bottom are larger networks. Bold is best (PSNR or SSIM) within its section. Blue is second best. As before, FLOPs calculated on 256x256 image. As TSANet code has not been released, we could not benchmark it on McMaster. For a similar reason, average PSNR/SSIM for TSANet does not include McMaster. *Binarized networks. } 
\label{table:hybridevs_results}
\end{table*}

\subsection{Real Image Unified Joint-Demosaicing-and-Denoising}

\begin{figure}
    \centering 
    \quad
    \includegraphics[width=0.47\textwidth]{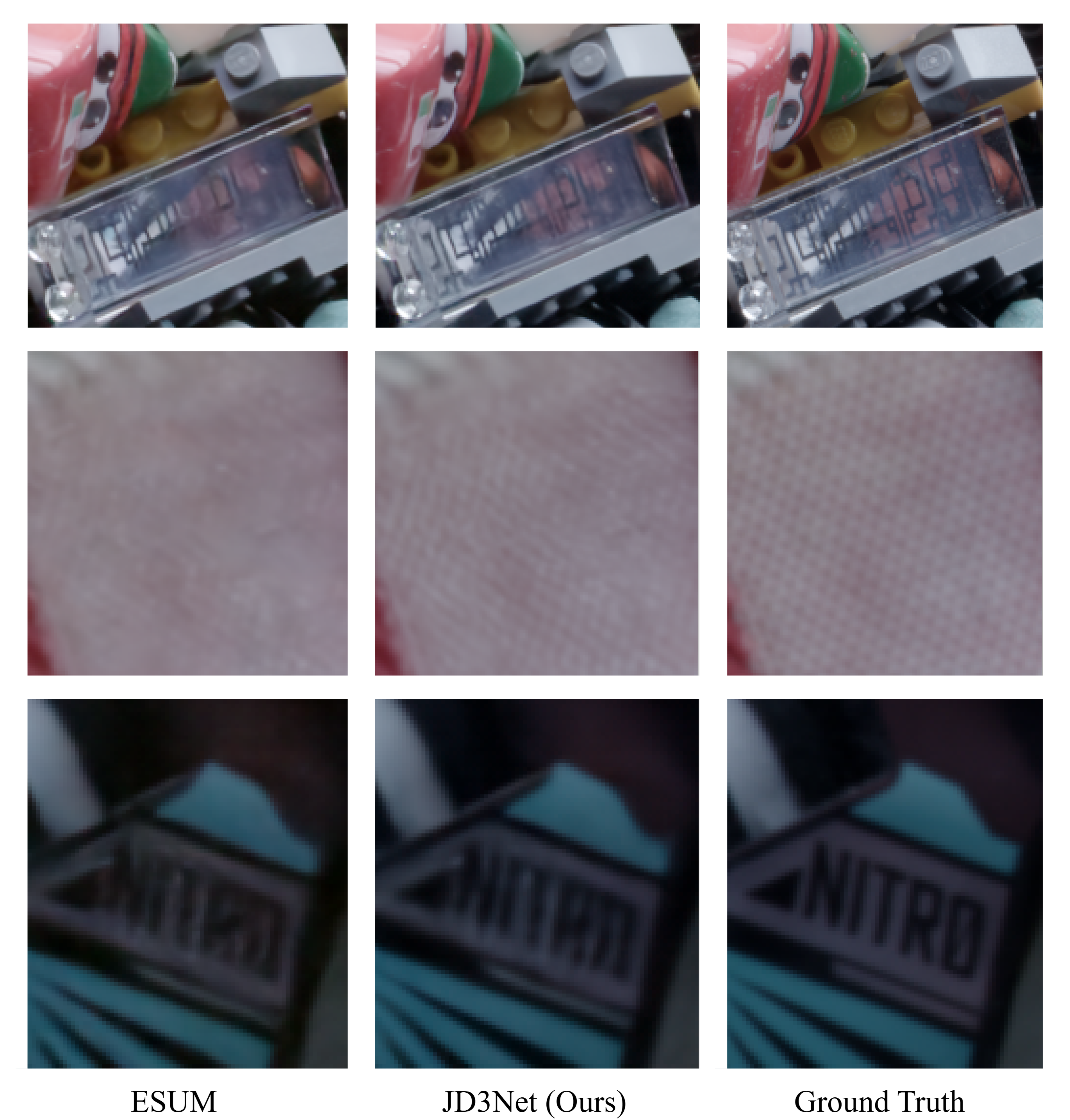}
    \caption{Qualitative Results for ESUM and JD3Net on HDD. All are ISO 3200 Nona Bayer. Demosaicing occurred on RAW images, but post-processing was done with a normal ISP pipeline to make the images visually normal. JD3Net produces more detailed and accurate images with less computational cost.}
    \label{fig:HDD_results} %
\end{figure}

To evaluate the performance of JD3Net and JD3Net-S comprehensively, we evaluate both JD3Net and JD3Net-S on all 4 difficulty settings of the Hard Demosaicing Dataset. Training settings are identical to those in \cref{sec:search}. In order to allow JD3Net to take multiple CFA patterns as input, we use ESUM's technique of appending CFAs to the input images. 

Averaged across all CFA patterns and all noise levels, JD3Net-S achieves only 0.1 PSNR less than ESUM\cite{ESUM} while being 16.3 times faster, while JD3Net outperforms ESUM by 0.35 PSNR while being 3.2 times faster. We show a selection of qualitative results between JD3Net and ESUM in \cref{fig:HDD_results}, which demonstrates the better reconstruction capability of JD3Net. Quantitative results are given in \cref{table:HDD_results}.

\subsection{Synthetic Quad-Bayer HybridEVS Demosaicing}

\begin{figure} 
    \centering 
    \quad
    \includegraphics[width=0.47\textwidth]{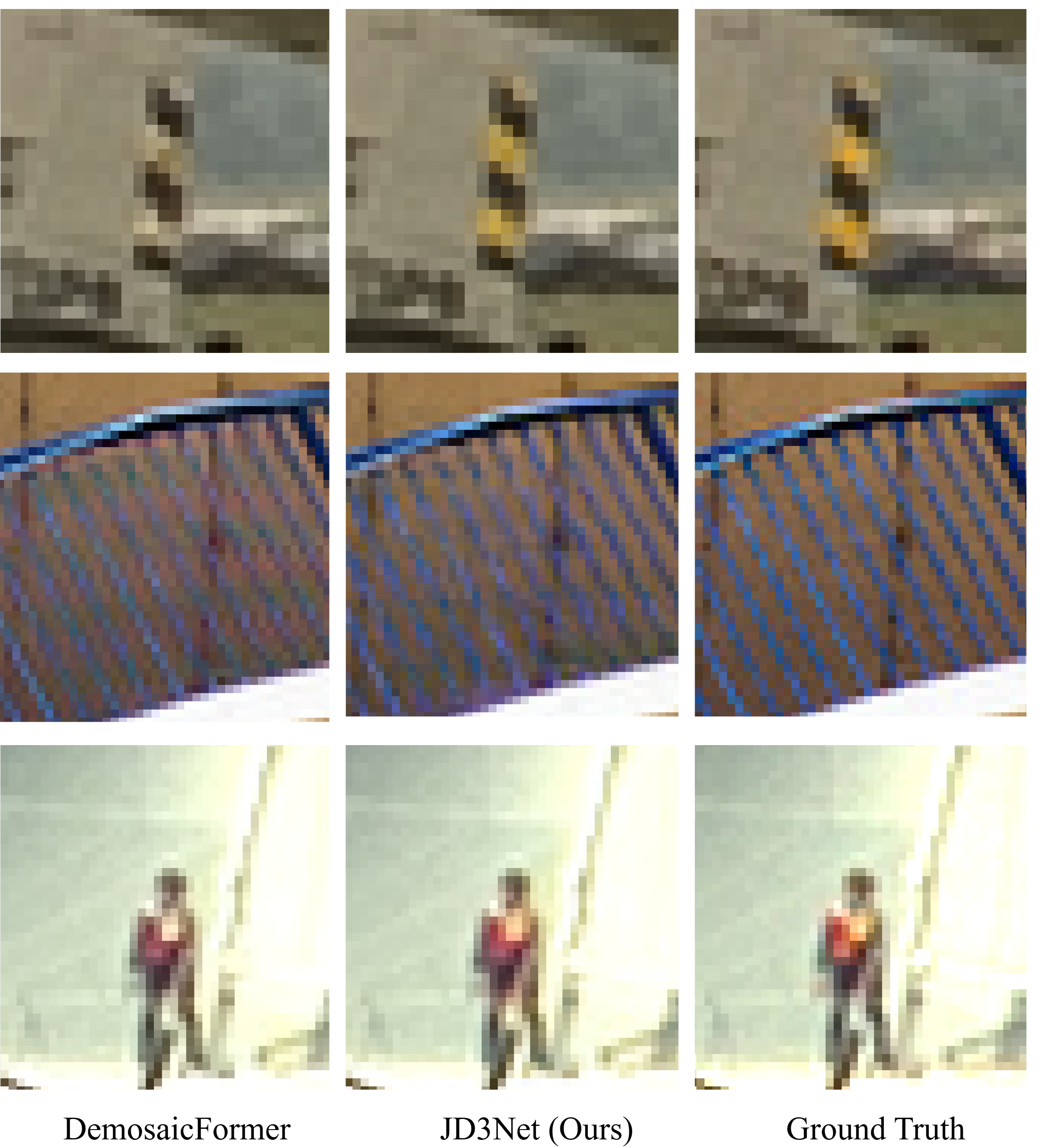}
    \caption{Qualitative Results for DemosaicFormer and JD3Net on Quad-Bayer HybridEVS demosaicing.}
    \label{fig:hybridevs_results} %
\end{figure}

We also evaluate JD3Net on synthetic quad-bayer HybridEVS demosaicing. We train on the MIPI dataset \cite{MIPI} training split for $8\cdot10^6$ iterations at a batch size of 20. We use PSNR loss\cite{NAFNet} and Adam optimizer with a learning rate of $0.001$. We apply the data augmentations used by DemosaicFormer\cite{DemosaicFormer} in the first part of their schedule, which is to synthetically generate rotated and flipped HybridEVS mosaics from MIPI's ground truth images. Following BMTNet\cite{BMTNet}, we evaluate on synthetically generated quad-bayer HybridEVS mosaics, which are generated from the Kodak\cite{Kodak}, McMaster\cite{McMaster}, BSD100\cite{BSD100} and Urban100\cite{Urban100} datasets, as well as the first hundred images of the Waterloo Exploration Database (WED)\cite{WED}. CFA appending is utilized for both JD3Net-S and JD3Net. 

Against DemosaicFormer\cite{DemosaicFormer}, the previous state-of-the-art method, JD3Net achieves comparable performance with 3.8 times fewer FLOPs, scoring better overall on Kodak, BSD100, and WED, and scoring better SSIM on McMaster and Urban100. Averaged across all 5 datasets, JD3Net improves by 0.17 PSNR compared to DemosaicFormer while being 3.8 times faster. JD3Net-S outperforms TSANet-S by 0.08 PSNR while being 48\% faster. Quantitative results are shown in \cref{table:hybridevs_results}. Some qualitative results of DemosaicFormer and JD3Net are shown in \cref{fig:hybridevs_results}, which demonstrates the improved reconstruction capability of JD3Net, in particular in reconstructing the original color.

\subsection{Synthetic Bayer Demosaicing}

We also evaluate JD3Net on synthetic bayer demosaicing with the Kodak\cite{Kodak} and McMaster\cite{McMaster} datasets. Following GRL\cite{GRL}, we train both networks on the ImageNet train split. JD3Net is trained on 128x128 patches at a batch size of 20, while JD3Net-S is trained on 129x129 patches (for divisibility by 3) at a batch size of 32, both for $10^6$ iterations. As before, we use PSNR loss and Adam optimizer with a learning rate of $0.001$, with random flipping and rotation augmentations. CFA appending is used for JD3Net-S, but since a 4x downsampling can learn the CFA pattern inherently, CFA appending is not necessary for JD3Net.

\begin{table}
\small	
\setlength{\tabcolsep}{3pt}
\renewcommand{\arraystretch}{1.5}
{\centering
\begin{tabular}{c|c|c|c}
\hline
Network & GFLOPs & Kodak & McMaster \\
\hline
DeepJoint\cite{DeepJoint} & 14 & 42.00 & 39.14 \\
RNAN\cite{RNAN} & 162 & 43.16 & 39.70 \\
DRUNet\cite{DPIR} & 278 & 42.68 & 39.39 \\
GRL-S\cite{GRL} & 491 & \color{blue} 43.57 & \textbf{40.22} \\
\hline
JD3Net-S & 25 & 42.27 & 39.44 \\
JD3Net & 128 & \textbf{43.65} & \color{blue} 40.07 \\
\hline
\end{tabular}
\par}
\caption{Kodak and McMaster Results (PSNR). Bold is best. Blue is second best. As before, FLOPs calculated on 256x256 image.} 
\end{table}

JD3Net achieves competitive performance with GRL-S while using 3.8 times fewer FLOPs, exceeding GRL-S's performance on Kodak but falling behind on McMaster. Additionally, JD3Net's architecure is notably much simpler than GRL, with JD3Net being fully convolutional and GRL employing complex attention mechanisms.

\subsection{Ablation}
\label{sec:ablation}

For the most part, ablation of JD3Net is performed implicitly through the process of its design. For instance, the effects of different depth/width ratios or downsampling can be found in \cref{sec:search}. Moreover, JD3Net is an extremely simple network; consequently, there are not many components to ablate on. Nonetheless, since we modify NAFNet's\cite{NAFNet} NAFBlock by removing attention, here we offer some brief ablations on the effect of attention on our networks, using the ISO3200 split of HDD. 

Attention ablation results can be found in \cref{table:ablation}. As expected, including NAFNet's Simple Channel Attention (SCA) improves the performance of JD3Net-S, but unexpectedly, it causes JD3Net to overfit and decreases performance. This can potentially be attributed to the relatively small size of real-image JDD datasets, and thus the fragility of network architectures trained upon them. Nonetheless, as the effects of attention are outside the topic of investigation of this paper, and JD3Net performs strongly without attention, we leave the design of attention for downsampled isotropic networks for future work.

\begin{table}
\small
\setlength{\tabcolsep}{3pt}
\renewcommand{\arraystretch}{1.5}
{\centering
\begin{tabular}{c|c|c|c}
\hline
Network & Single & Quad & Nona \\
\hline
JD3Net-S & 49.06 & 48.61 & 48.26 \\ 
JD3Net-S + SCA & 49.10 & 48.65 & 48.30 \\
\hline
JD3Net & 49.36 & 48.99 & 48.54 \\
JD3Net + SCA & 48.95 & 48.54 & 48.06 \\
\hline
\end{tabular}
\par}
\caption{Ablation testing for channel attention in JD3Net. SCA is simple channel attention. Results are test PSNR for ISO3200 HDD. Channel attention improves JD3Net-S but causes JD3Net to overfit.} 
\label{table:ablation}
\end{table}

%% file: sec/5_conclusion.tex
\section{Conclusion}

This paper has shown the effectiveness of downsampling in isotropic networks for JDD and image demosaicing on a variety of datasets. JD3Net significantly exceeds both state-of-the-art performance and efficiency for JDD on the Hard Demosaicing Dataset. For Bayer Demosaicing and Quad-Bayer HybridEVS Demosaicing, JD3Net achieves competitive performance and state-of-the-art efficiency. Notably, JD3Net is a simple fully convolutional network, while many of the methods we compared against in this paper, such as GRL or DemosaicFormer, employ substantially more complex and custom-tuned mechanisms, such as custom attention or state-space modules. Finally, while the focus of this paper was isotropic networks, similarly JD3Net outperformed many strong U-Net architectures. Overall, JD3Net is a simple demonstration of the effectiveness of downsampling for JDD, and we expect that future isotropic networks can leverage both spatial downsampling and special JDD/demosaicing-specific network design to achieve even greater performance. 